\documentclass[conference]{IEEEtran}
\IEEEoverridecommandlockouts
\usepackage{enumitem}
\usepackage{hyperref}
\usepackage{cite}
\usepackage{amsmath,amssymb,amsfonts}
\usepackage{algorithmic}
\usepackage{graphicx}
\usepackage{textcomp}
\usepackage{xcolor}
\usepackage{graphicx}
\usepackage{tcolorbox}
\usepackage{amsmath}
\usepackage{tcolorbox}
\usepackage{multicol}
\usepackage{subcaption} 
\usepackage{cite}
\usepackage{amsmath,amssymb,amsfonts}
\usepackage{algorithmic}
\usepackage{textcomp}
\usepackage{xcolor}
\usepackage{stfloats}      
\usepackage{graphicx}
\usepackage{subcaption}
\usepackage{tcolorbox}
\usepackage{booktabs, xcolor, colortbl}  
\definecolor{rowcollight}{gray}{0.95}    
\tcbuselibrary{listings, breakable}
\lstdefinelanguage{json}{
    morestring=[b]",
    morecomment=[l]{//},
    morekeywords={true,false,null},
    sensitive=false,
}

\lstdefinestyle{jsonStyle}{
    language=json,
    basicstyle=\footnotesize\ttfamily,
    breaklines=true,
    showstringspaces=false,
    backgroundcolor=\color{gray!5},
    frame=single
}

\def\BibTeX{{\rm B\kern-.05em{\sc i\kern-.025em b}\kern-.08em
    T\kern-.1667em\lower.7ex\hbox{E}\kern-.125emX}}
\begin{document}

\title{Agentic AI framework for End-to-End Medical Data
Inference}

\author{

\IEEEauthorblockN{Soorya Ram Shimgekar}
\IEEEauthorblockA{\textit{University of Illinois - Urbana Champaign} \\
sooryas2@illinois.edu}
\and

\IEEEauthorblockN{Shayan Vassef}
\IEEEauthorblockA{\textit{University of Illinois - Chicago} \\
svass@uic.edu}
\and

\IEEEauthorblockN{Abhay Goyal}
\IEEEauthorblockA{\textit{Missouri S\&T} \\
aghnw@umsystem.edu}
\and

\IEEEauthorblockN{Navin Kumar}
\IEEEauthorblockA{\textit{Nimblemind.ai} \\
navin@nimblemind.ai}
\and 

\IEEEauthorblockN{Koustuv Saha}
\IEEEauthorblockA{\textit{University of Illinois - Urbana Champaign} \\
ksaha2@illinois.edu}
}

\maketitle

\begin{abstract}
Building and deploying machine learning solutions in healthcare remains expensive and labor-intensive due to fragmented preprocessing workflows, model compatibility issues, and stringent data privacy constraints. In this work, we introduce an Agentic AI framework that automates the entire clinical data pipeline, from ingestion to inference, through a system of modular, task-specific agents. These agents are capable of handling both structured and unstructured data, enabling automatic feature selection, model selection, and preprocessing recommendation without manual intervention.
We evaluate the system on publicly available datasets from geriatrics, palliative care, and colonoscopy imaging. For example, in the case of structured data (anxiety data) and unstructured data (colonoscopy polyps data), the pipeline begins with file-type detection by the ``Ingestion Identifier Agent", followed by the ``Data Anonymizer Agent" ensuring privacy compliance, where we first identify what type of data it is and then anonymize it. The ``Feature Extraction Agent" then identifies features using an embedding-based approach for tabular data, which gives us all the column names, and a multi-stage MedGemma-based approach for image data, which gives us the modality and disease name. These features guide the ``Model-Data Feature Matcher Agent" in selecting the best-fit model from a curated repository. The ``Preprocessing Recommender Agent" and ``Preprocessing Implementor Agent" then apply tailored preprocessing based on data type and model requirements. Finally, the ``Model Inference Agent" runs the selected model on the user uploaded data and generates interpretable outputs using tools like SHAP, LIME, and DETR attention maps.
By automating these high-friction stages of the ML lifecycle, the proposed framework reduces the need for repeated expert intervention, offering a scalable and cost-efficient pathway for operationalizing AI in clinical environments.

\end{abstract}


\section{Introduction}

The integration of Artificial Intelligence (AI) into clinical workflows holds transformative potential for healthcare, enabling timely, data-driven decision-making across diagnosis and treatment planning \cite{topol2019deep}. 
Despite advances in AI infrastructure, deploying machine learning models from raw clinical data remains fragmented and highly manual. Up to 80 percent of a data scientist’s time is spent on tasks like data preprocessing, model selection, and pipeline setup, rather than on model development and evaluation \cite{schmetz2024inside}. These workflows often require large interdisciplinary teams including clinicians, data engineers, machine learning experts, and privacy officers \cite{nair2024comprehensive}, costing healthcare institutions between \$850,000 and \$1.5 million annually \cite{mckinsey, aalpha}, \cite{sanfran}. This reliance introduces delays, human error, and financial strain, particularly for institutions handling large volumes of patient data \cite{hassan2024barriers}.

Beyond cost and scalability, clinical AI adoption faces additional challenges related to privacy, model-data alignment, and data heterogeneity \cite{rieke2020future, subbaswamy2020preventing}, \cite{beam2018big, miotto2017deep}. Legal frameworks such as the Health Insurance Portability and Accountability Act (HIPAA)\cite{hipaa} and the General Data Protection Regulation (GDPR)\cite{gdpr} mandate strict protection of personal health information, requiring AI systems to integrate anonymization and explainability as default capabilities rather than add-ons \cite{marks2023ai}. Likewise, model-data mismatch, selecting inappropriate models for the characteristics of available data, can severely impact performance or reliability \cite{he2021automl}. Yet model selection remains a domain-specific, manual task that does not scale in high-volume or time-sensitive environments. Further complicating matters, clinical data is often multimodal, high-dimensional, and sparsely labeled. Real-world deployment demands infrastructure that can automatically clean, interpret, and standardize diverse data types, a problem that remains unresolved despite progress in model architectures \cite{isern2010agents}.

A promising strategy for addressing these systemic challenges is Agentic AI, a paradigm that defines AI systems as collections of autonomous, modular ``agents" each assigned specific roles and goals \cite{acharya2025agentic}. By distributing tasks among specialized agents, the system becomes more flexible, interpretable, and efficient in handling complex clinical workflows. These agents can independently perceive, reason, act, and communicate, allowing them to operate collaboratively or autonomously depending on context \cite{acharya2025agentic}.

Accordingly, we introduce an Agentic AI framework that modularly orchestrates clinical workflows, including data ingestion, anonymization, model training, inference, and explanation, through a network of domain-specialized agents. Together, these agents form an end-to-end system capable of autonomously translating raw, multimodal clinical data into privacy-preserving, interpretable predictions.
 


\section{Literature Review}

Agentic and modular architectures are increasingly being explored as alternatives to rigid end-to-end systems, particularly in domains that require interpretability and adaptability \cite{masterman2024landscape}. 


In clinical AI, agentic designs are especially relevant due to the complexity and variability of medical data pipelines. Traditional pipelines tend to be vertically integrated—focusing on optimizing single-model performance—whereas agentic frameworks support horizontal integration by coordinating multiple tasks that span the full data-to-deployment lifecycle. Language models acting as agents that can reason, maintain context, and chain tasks have shown promising results in domains such as medical question answering \cite{lievin2024can, singhal2023large}, yet broader adoption in operational clinical systems remains limited.

In an agentic framework, each critical subtask in clinical AI can be executed by a specialized, intelligent agent. Here are some examples. 


\textbf{Preprocessing Agents.}
Preprocessing is a fundamental step in clinical machine learning pipelines, involving transformations such as imputation, normalization, encoding, temporal alignment, and feature engineering—all of which critically impact model performance and interpretability \cite{almuhaideb2016individualized}.

Although AutoML frameworks like AutoGluon \cite{agtabular} and AutoKeras \cite{jin2019auto} automate basic preprocessing, they often lack the contextual sensitivity required for clinical data. In particular, they may overlook best practices such as performing imputation or scaling within cross-validation folds.

Agentic AI systems offer a more adaptive solution by dynamically orchestrating preprocessing steps based on task requirements, data characteristics, and domain constraints. Recent systems such as ELT-Bench \cite{jin2025elt} and Bel Esprit \cite{kim2024bel} showcase the potential of LLM-based agents to autonomously construct end-to-end pipelines, including transformation and feature processing logic. ELT-Bench provides a benchmark to assess the capability of agents to generate Extract-Load-Transform (ELT) pipelines using LLMs that interpret metadata and data schema to configure pipelines automatically \cite{jin2025elt}. Similarly, Bel Esprit offers a multi-agent conversational framework where agents collaboratively refine a pipeline from user intent to model deployment, often engaging in preprocessing selection based on context \cite{kim2024bel}.

In the context of scalable and distributed data processing, frameworks such as Intelligent Spark Agents demonstrate how LLMs can orchestrate Spark SQL and DataFrame APIs to perform preprocessing and transformation tasks in a modular, adaptive fashion \cite{wang2024intelligent}. These agents dynamically alter their behavior based on data types and real-time feedback from their environment, enabling context-aware feature transformations and more efficient processing for streaming or batch clinical data. Other systems, such as GoldMiner, adopt a distributed “data worker” model to scale stateless preprocessing operations (like imputation and normalization) across clusters \cite{zhao2023goldminer}.

\textbf{Privacy and Compliance Agents.}
Agent-based privacy frameworks offer a dynamic, context-aware approach: they can detect, redact, or anonymize sensitive data at multiple stages of the pipeline—from ingestion through model output. One example is a HIPAA‑compliant agentic AI framework that integrates attribute-based access control (ABAC), hybrid PHI sanitization, and immutable audit logging. This system employs structured rule‑based detection (e.g., regex for SSNs or medical record numbers), combined with a BERT-based model fine-tuned on clinical corpora to spot contextual PHI in unstructured notes \cite{neupane2025towards}. These agents make context-sensitive redaction decisions and enforce policies such as “minimum necessary” access while preserving audit trails that align with regulatory mandates like HIPAA and GDPR.

For structured and unstructured data alike, modern agents leverage LLMs to perform high-precision anonymization. The LLM‑Anonymizer pipeline benchmarks Llama‑2/3 based models on real-world clinical documents, achieving a recall above 97\% and a precision around 99\%—comparable to GPT‑4 \cite{wiest2024anonymizing}. It supports flexible entity definitions, OCR processing for scanned records, and in-context prompt refinement, making it a robust tool for document-level PII redaction without cloud dependencies or data exfiltration.

Beyond redaction, advanced privacy agents can employ cryptographic and decentralized techniques such as federated learning (FL), differential privacy (DP), and homomorphic encryption (HE). A multisite federated learning framework, enhanced with differential privacy, was shown to train predictive models over one million real-world EHR records without transferring raw data, while preserving accuracy and offering provable privacy guarantees \cite{choudhury2019differential}.

Moreover, commercial-grade PII detection agents such as NuExtract (numind/NuExtract-v1.5) exhibit high accuracy across diverse document types, outperforming standard NER pipelines \cite{numind_nuextract15}. These agents intelligently distinguish PII from other content, maintain entity consistency across documents via deterministic mapping, and preserve format and context by performing context-aware replacement (e.g., gender consistent synthetic names), all validated through system-level output checks and optional human review workflows.

\textbf{Feature Matching and Model Selection Agents.}
Selecting models that align with both the statistical structure and semantic content of clinical data is a complex and error-prone task—one that model selection agents aim to automate intelligently. These agents analyze input datasets in terms of modality (e.g., text, tabular, imaging), dimensionality, schema structure, and domain-specific semantics. They then match them against repositories of pretrained or fine-tunable models, ensuring compatibility before deployment.

A key strategy employed by such agents is retrieval-augmented generation (RAG) combined with embedding-based similarity matching. For example, the LLM-Match framework uses RAG to link patient records with eligibility criteria in clinical trials, fine-tunes models, and outperforms zero-shot GPT-4 baselines on multiple datasets \cite{li2025llm}. Another approach is OBOE (AutoML via Collaborative Filtering), which formulates model selection as a matrix completion problem using performance data across multiple models and datasets \cite{yang2019oboe}. By leveraging fast landmarkers and a low-rank approximation of performance matrices, OBOE embeds both datasets and models in a shared latent space, allowing agents to recommend high-performing models based on meta-feature similarity. 
Complementing these, feature-based meta-learning frameworks use handcrafted and learned meta-features to guide model choice. 
A recent study demonstrated how autoencoder-based meta feature extraction improves model selection by learning abstract representations of dataset complexity and distribution \cite{garouani2023autoencoder}. 
Similarly, in \cite{peng2019moment}, the authors introduced a moment-matching autoencoder that aligns input distributions between new and historical datasets to enhance compatibility estimation between models and data. These feature-alignment techniques support agents in identifying models that generalize well across dataset shifts—a common challenge in real-world healthcare data.

\begin{table*}[t]
\centering
\sffamily
\renewcommand{\arraystretch}{1.5}  
\setlength{\tabcolsep}{1pt}
\begin{tabular}{p{0.4\textwidth}p{0.6\textwidth}}

\toprule
\rowcolor{rowcollight}
\textbf{Agent} & \textbf{Purpose and Functionality} \\
\midrule

Feature Identifier Agent \newline \texttt{[Ingestion\_Classifier]} &
Detects and classifies file types (e.g., CSV, Excel, ZIP) using Magika, so data-specific downstream workflows can be used.\\

\rowcolor{rowcollight}
Data Anonymization Agent \newline \texttt{[Ingestion\_Anonymizer]} &
Automatically detects and redacts PII from both structured(tabular) and unstructured(image) data using Google Cloud DLP. \\

Feature Extraction Agent \newline \texttt{[Ingestion\_Selector]} &
Extracts semantic ``headers" for both tabular (column names) and image data (Modality, Disease Type). \\

\rowcolor{rowcollight}
Model-Data Matcher Agent \newline \texttt{[Ingestion\_Feature\_Matcher]} &
Matches user data to the most suitable AI model, using the ``headers" based on the model database \autoref{table:json}. Ensures model-data compatibility. \\

Preprocessing Recommender Agent \newline \texttt{[Preprocessing\_Recommender]} &
Recommends preprocessing operations for both structured(tabular) and unstructured(image) data based on the identified ``headers". Supports automated or custom user-defined preprocessing based on the data size. \\

\rowcolor{rowcollight}
Preprocessing Implementor Agent \newline \texttt{[Preprocessing\_Implementor]} &
Executes the preprocessing pipeline suggested by the recommender agent. \\

Model Inference Agent \newline \texttt{[Model\_Inferencer]} &
Runs selected models for final prediction. Supports interpretability via SHAP, LIME, and attention visualizations for respective modalities. \\

\bottomrule
\end{tabular}
\caption{Autonomous AI Agents in the Clinical Data Pipeline and Their Roles}
\label{table:agents_summary}
\end{table*}

\textbf{Orchestration and Explanation Agents.}
In complex AI systems, particularly in clinical domains, orchestrating multiple interdependent components, such as data preprocessing, model selection, privacy filtering, and interpretation, requires modular yet cohesive coordination. This task of managing the workflow of various interdependent agents is performed by an orchestration agent. Various strategies are proposed to get optimal behaviour from these agents, such as in \cite{chen2023agentverse}, where the authors introduced, AgentVerse, a multi-agent LLM framework where a central orchestrator governs task flows among collaborative agents for data processing, modeling, and communication, demonstrating high modularity and scalability in autonomous systems. 

Along with orchestrating the workflow, it is also important to understand the thought process behind the end output, which requires a level of transparency and explainability. These explanations require the agents to have some human-readable reasoning behind preprocessing justifications, data lineage, model rationale, or predictive reasoning. In \cite{lu2024causal}, Causal State Distillation was introduced, which breaks down agent decision-making into causally salient reward components, furthering interpretability in reinforcement learning contexts. 
Basappa et al. proposed a system where agents perform self-explanation using introspective reasoning and structured prompt chaining. These agents generate natural language justifications for their actions and beliefs by reflecting on their internal state—a capability validated in educational AI systems to enhance transparency and accountability \cite{basappa2025self}. Similarly, the Agent-Based Explanations framework underscores the importance of communicative protocols in multi-agent explanation, ensuring that explanations are tailored to the explainee’s knowledge level and context, especially important in clinician-facing environments \cite{ciatto2020agent}.

\section{Data}
Data utilized in this study was obtained from publicly available sources. For models focused on geriatric applications—such as fall prediction—we employed the GSTRIDE \cite{garcia2022gstride} dataset, which comprises multimodal sensor data including foot pressure, foot angle, and ankle related features. These inputs allow the models to effectively capture fine-grained details of an individual’s gait and movement dynamics. Our work in the case of geriatrics was to predict if the patient would fall based on the different numeric and other information we can get.  

For the palliative care-focused hope prediction model, we used data \cite{mano2018classification},  \cite{hofmann2017patients} from an open-access dataset. The data consisted of different features such as the age, gender, consumption of antidepressants, vomiting etc. This model aims to predict levels of hope in patients receiving palliative care, supporting clinical decision-making and psychological assessment. 

For polyp‐type classification and bounding‐box prediction, we used the dataset introduced in \cite{li2021colonoscopy}, which provides an annotated compilation of three publicly available sources: the CVC‐ColonDB dataset \cite{bernal2012towards}, the GLRC dataset \cite{mesejo2016computer}, and the KUMC dataset, comprising 80 colonoscopy video sequences from the University of Kansas Medical Center.

\section{Agent pipeline}

\begin{figure*}[t]
  \centering
  \includegraphics[width=0.9\textwidth]{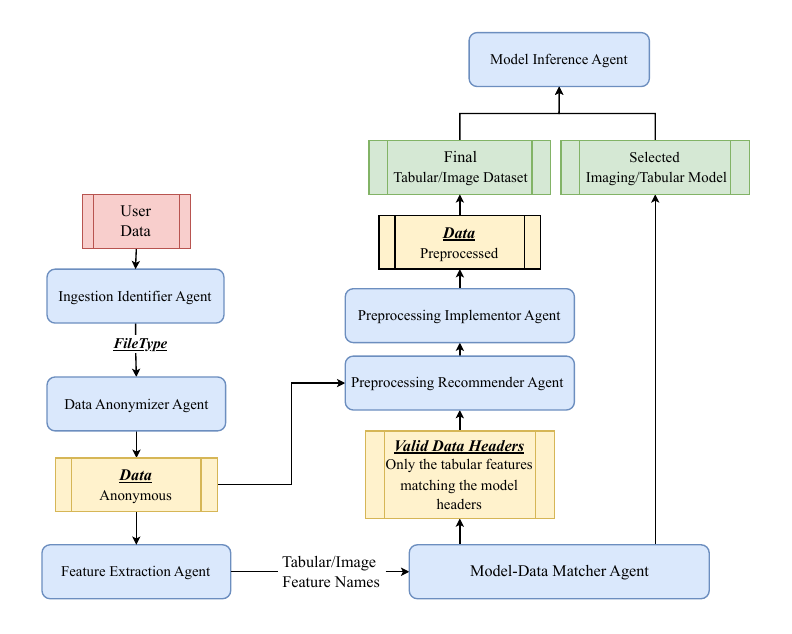}
  \caption{Complete architecture}
  \label{fig:arch}
\end{figure*}


\begin{figure*}[t]
\begin{tcolorbox}[
  colback=gray!5,
  colframe=gray!40,
  title=Simplified Model Database (JSON),
  listing only,
  listing options={style=jsonStyle}
]
\begin{lstlisting}
{
  "table": {
    "MODEL_01": {
      "modality": "anxiety prediction",
      "headers": ["age", "gender", "ECOG", "living_situation" ..... "anxiety"],
      "output": "anxiety"
    },
    ...
  },
  "image": {
    ...
    "MODEL_02": {
      "modality": "colon colonoscopy scan",
      "caption": "Detects and classifies hyperplastic vs. adenomatous polyps in
      colonoscopy images"
    },
    ...
  }
}
\end{lstlisting}
\end{tcolorbox}
\caption{Model Database}
\label{table:json}
\end{figure*}

We introduce autonomous AI agents designed to operate within a clinical data pipeline, from enrichment to inference. These agents are tailored to support healthcare providers in transforming raw multimodal clinical data into interpretable insights. Each agent is characterized by clinical reasoning, contextual awareness, and alignment. Agents operate across various stages, including ingestion, model selection, preprocessing, modeling, and explainability. The complete architecture with all the agents interacting with each other and the data is seen in \autoref{fig:arch}. A summary of all the AI agents along with their role is seen in \autoref{table:agents_summary}. The entire workflow is orchestrated using Google's Agent Development Kit (ADK) \cite{googleADK2025}. Below we present all our agents that complete the above-stated tasks. \newline

\textbf{Ingestion Identifier Agent: } 
The Feature Identifier Agent serves as the initial component in the clinical data pipeline, responsible for classifying uploaded files into recognized data formats to enable context-aware downstream processing. It supports various formats—including \texttt{CSV}, \texttt{XLSX}, \texttt{JSON}, and \texttt{ZIP} and employs Magika \cite{fratantonio2024magika}, a deep learning-based MIME-type detection framework to analyze the uploaded data's structural and byte-level content for filetype detection.

For example, when detecting an Excel file, it verifies compliance with the Open XML Spreadsheet standard and returns a corresponding MIME string (e.g., \texttt{application/vnd.openxmlformats-officedocum ent.spreadsheetml.sheet}).

For compressed archives like \texttt{ZIP}, the agent performs in-memory recursive unpacking, navigating nested directory structures to access the innermost files. Each extracted file is individually classified into a MIME string, allowing detection of multiple formats within a single archive. The agent then produces a structured summary of all the uniquely identified file types.

By automating file-type detection, the Feature Identifier Agent reduces manual preprocessing and ensures that workflows corresponding to the correct data type are used in the downstream processes. \newline


\textbf{Data Anonymizer Agent: }
The Data Anonymization Agent ensures compliance with data privacy and governance standards by performing automated detection and masking of personally identifiable information (PII). This agent supports both structured and unstructured data modalities and leverages the Google Cloud Data Loss Prevention (DLP) API for entity-level anonymization \cite{google_dlp_doc}.

For structured data (e.g., CSV, Excel), the agent applies DLP’s inspection engine to scan for a comprehensive set of PII types, including but not limited to: \textit{names, email addresses, phone numbers, medical record numbers, credit card numbers, IP addresses, and dates of birth}. Identified elements are then masked with fixed-length placeholder tokens (e.g., \texttt{"****"}), preserving schema integrity while ensuring content obfuscation.

For image data, the agent utilizes DLP’s visual inspection capabilities to detect PII embedded in the image, such as textual identifiers. These regions are redacted using opaque overlays (e.g., black rectangles), rendering sensitive content unreadable.

By integrating both text and visual anonymization into the ingestion pipeline, the agent enables privacy-preserving transformations across heterogeneous clinical data. This design ensures alignment with regulatory frameworks (e.g., HIPAA, GDPR) and establishes a secure foundation for further data preprocessing and modeling tasks. \newline

\textbf{Feature Extraction Agent: }
The Feature Extraction Agent performs automated, modality-specific feature identification, which is crucial for understanding what the data is about. By adapting its logic to the input format, the agent ensures that relevant semantic metadata is extracted.

For structured data, the agent uses column names as proxies for feature descriptors, referred to from here on as ``headers". These headers are treated as candidate variables for subsequent model selection, preprocessing, and modeling tasks.

For unstructured image data, our framework integrates MedGemma, a medical vision–language model developed by Google \cite{lin2025medgemma}. A random anonymized image from the previous step is passed through a multi-stage classification pipeline to return image data specific ``headers", namely, (1) Modality and (2) Disease-Type. The agent first classifies the imaging ``Modality", such as scan, ``breast histopathology scan", ``colon colonoscopy scan", etx, by matching visual features against a curated set of modality descriptions, shown in the model database \autoref{table:json}. Following modality recognition, we further prompt MedGemma to classify the most likely ``Disease Category" present in the image, based on the detected Modality, and ``Caption" of the corresponding Modality from the model database \autoref{table:json}. The classification at each step is governed by confidence thresholds and decision rules, allowing the agent to handle uncertain inputs by either terminating or requesting re-analysis of a new random anonymized image. Together, these two levels of semantic extraction(Modality and Disease) enable routing of image data into specialized diagnostic pipelines and inform the selection of the most appropriate AI model. 

\begin{figure}[t]
  \centering
  \includegraphics[width=0.5\textwidth]{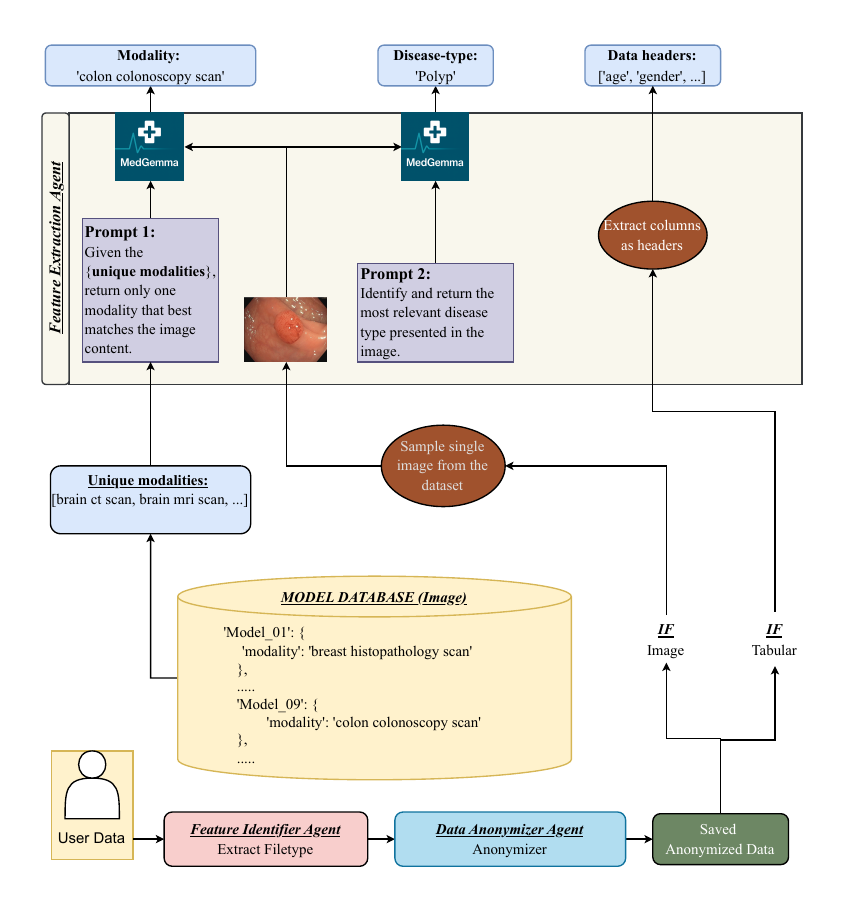}
  \caption{Ingestion Selector framework}
  \label{fig:ingest}
\end{figure}

The interaction among this agent and upstream agents, such as the Feature Identifier Agent and Data Anonymization Agent, is illustrated in \autoref{fig:ingest}. \newline

\textbf{Model-Data Matcher Agent: }

\begin{figure*}[t]
  \centering
  \includegraphics[width=1.03\textwidth]{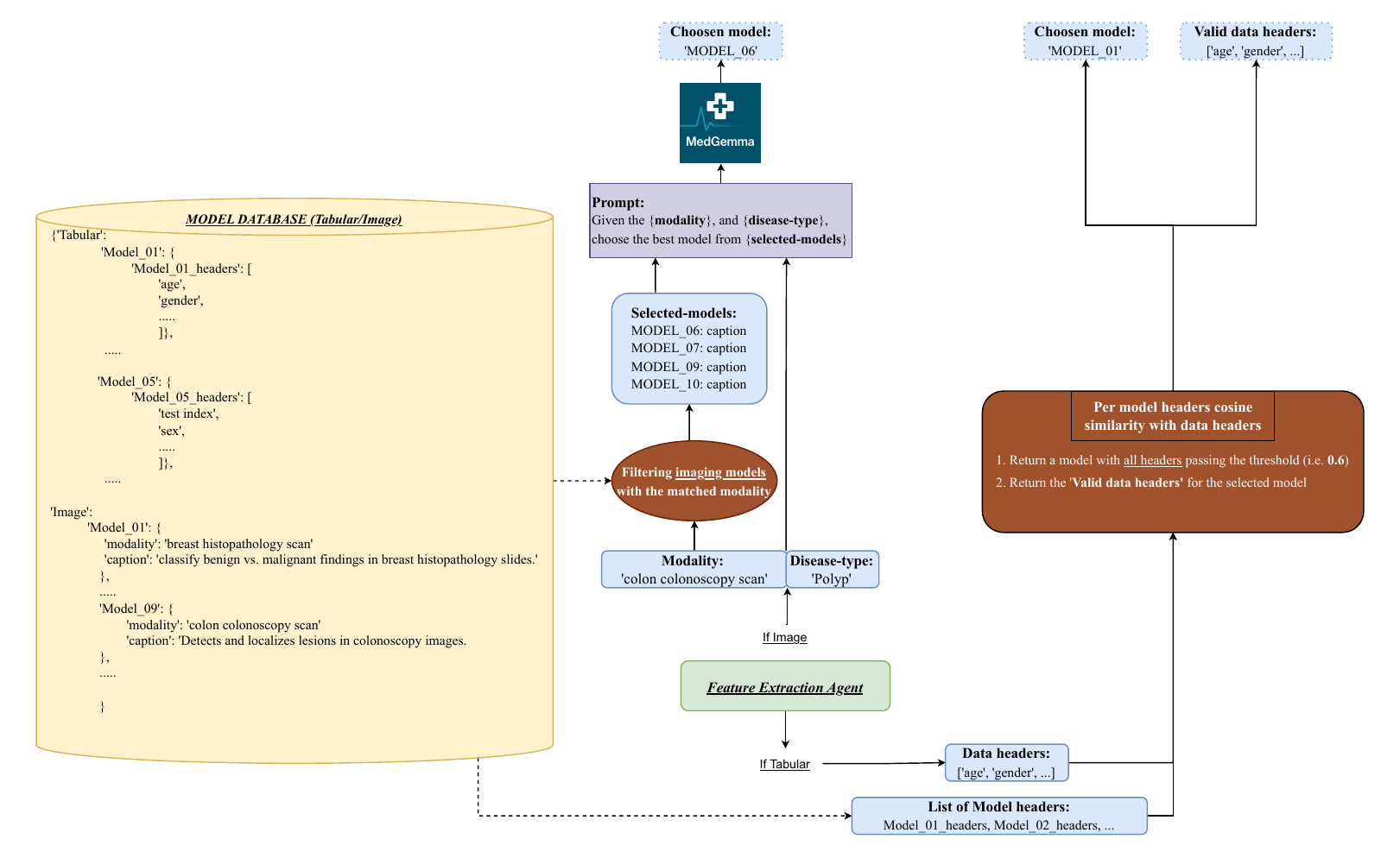}
  \caption{Ingestion Feature Matcher framework}
  \label{fig:rag}
\end{figure*}

The Model-Data Feature Matcher agent bridges the gap between raw data ingestion and model deployment by selecting the most appropriate model from a curated repository, based on semantic alignment between input features and model requirements.

For structured (tabular) data, the agent operates by comparing the column headers of the user uploaded dataset against the required headers of each candidate model stored in the model database \autoref{table:json}. To enable robust semantic comparisons between the headers of user data and candidate models, each header is first transformed into a fixed-length embedding vector (size=768) using the SapBERT model \cite{liu2020self}, which is pretrained on biomedical literature and excels at encoding medical terminology. The comparison process uses cosine similarity to evaluate semantic proximity between the uploaded user dataset and saved model headers. A model is deemed eligible if every required feature can be matched to a dataset column with a similarity score exceeding a threshold (empirically set to 0.6). 

To prevent redundant matches, the agent ensures that each user uploaded dataset column is assigned to at most one required header, using a greedy selection scheme that iteratively picks the best available match for each required field. If all required headers for a given model are successfully matched, the agent selects that model as a candidate. Among all such candidates, the LLM-powered agent then chooses the best model based on the model description. The final output includes the best model name and a filtered dataset containing only the headers required by the selected model, ensuring compatibility and preventing downstream schema mismatches with the selected model.

In contrast, for unstructured image data, the agent leverages the MedGemma vision–language model \cite{lin2025medgemma} to use the image ``Modality" and ``Disease Type" to select the model that best aligns with both the modality and the identified disease.

\autoref{fig:rag} illustrates this architecture, highlighting the decision flow between the Feature Extraction Agent, the MedGemma Agent (for image inputs), and the Model Matching Agent. \newline


\textbf{Preprocessing Recommender Agent, Preprocessing Implementor Agent: }

The Preprocessing Recommender Agent autonomously suggests optimal preprocessing strategies that are tailored to both the structure of the uploaded dataset and the specific requirements of the selected machine learning model. This agent supports both user-guided and fully automated modes: when the dataset size exceeds a threshold (e.g., 50 MB), user-specified selection of preprocessing steps is disabled, and is automatically selected based on the selected model requirements.

For tabular data, the agent begins by extracting metadata for each header, capturing attributes such as column name, data type, number of null and unique values, and minimum, maximum values. Based on this metadata, it infers the column type, classifying each column as ``Binary", ``Categorical", ``Numerical", or ``Textual". This classification is performed using a rule-based heuristic: binary columns are identified by the presence of exactly two unique values; categorical columns are determined by low unique values having short string lengths; numerical columns are identified by data type and value distribution with the number of unique values exceeding 0.8*(total rows of the dataset); and columns that do not fit these criteria are labeled as text. This labeling guides the selection of preprocessing steps for each header. Based on inferred column types, the agent recommends standard preprocessing strategies to the user.

In the case of image data, preprocessing is not manually configured by the user. Instead, the system employs a model-specific preprocessing pipeline that is tightly coupled with the selected image model. This design choice reflects a growing trend in modern vision architectures, such as the DEtection TRansformer (DETR) and its variants, where preprocessing routines are not generic but are instead co-trained with the model during its original development. These pipelines may include learned resizing strategies, custom normalization schemes, or image tokenization methods optimized for the model’s attention mechanism and feature extraction layers. This approach prevents mismatch errors inherent in manual preprocessing and enhances modularity by allowing each model to encapsulate its own preprocessing logic, thereby streamlining deployment across diverse image-based tasks.

Complementing the Preprocessing Recommender Agent, the Preprocessing Implementor Agent is responsible for applying the selected preprocessing steps to the dataset. It takes the anonymized, feature-matched data from the previous steps and executes each preprocessing step given by the preprocessing recommender.

\begin{figure*}[t]
  \centering
  \includegraphics[width=\textwidth]{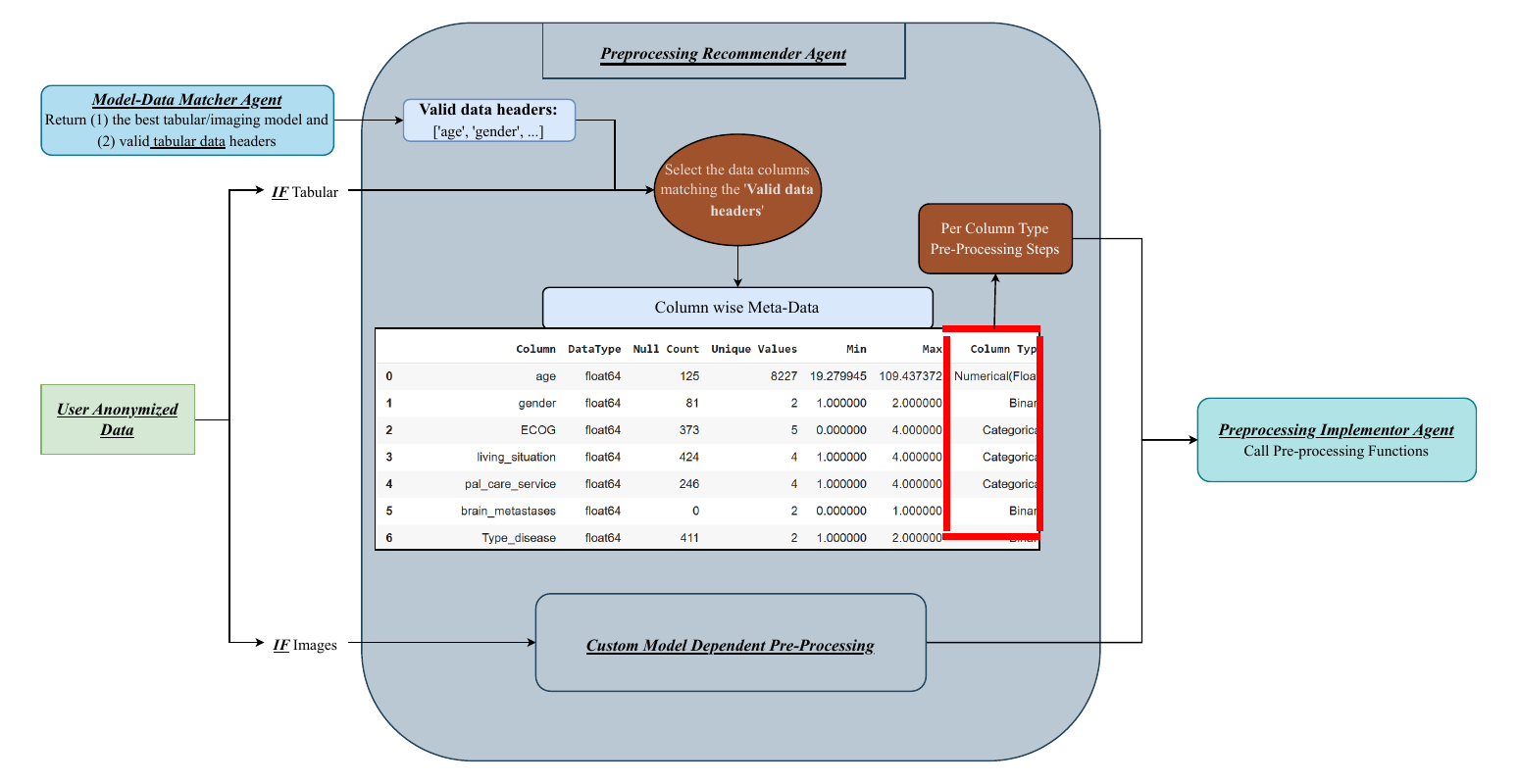}
  \caption{Preprocessing framework}
  \label{fig:prep}
\end{figure*}

The interaction between both agents and their integration with the previous agents is illustrated in \autoref{fig:prep}. \newline


\textbf{Model Inference Agent: }

The final stage of the pipeline is handled by the Model Inference Agent, which is responsible for applying trained machine learning models to the processed input data and generating interpretable outputs. This agent supports both tabular and image-based modalities, which then invokes specialized modeling architectures that are optimized for each data type requirement.

If the size of data is small, we use a gradient boosting machine (GBM) type of algorithm referred to as Perpetual Booster  \cite{marasco2025enhancing}, and in the case of larger-sized data, we use a custom deep learning architecture. These models are trained using the anonymized preprocessed data generated earlier in the pipeline and are then applied to either the held-out test set or newly uploaded, unseen data. The final model outputs are compiled into a structured CSV file. Additionally, the agent supports model interpretability through integration with established explainability tools such as SHapley Additive exPlanations (SHAP)\cite{lundberg2017unified} and Local Interpretable Model-agnostic Explanations (LIME) \cite{biecek2021local}. When requested, the agent computes these explanations and identifies the top 10 most influential features based on the selected method. These features can optionally be used to retrain/finetune the model on a reduced feature set, further optimizing and enhancing the robustness of the model.

For image data, inference is performed using a fine-tuned variant of the DEtection Transformer (DETR) architecture \cite{carion2020end}, specifically customized for clinical vision tasks. The specific saved model is selected based on the imaging modality and disease type extracted earlier. For instance, when processing colonoscopy images associated with colon polyps, the agent retrieves a DETR model fine-tuned to detect and classify hyperplastic versus adenomatous polyps. The model produces bounding box predictions overlaid directly on the original image to indicate regions of clinical interest. In parallel, a structured CSV file is generated to document bounding box coordinates and class labels, providing a tabulated summary of all visual predictions.

To enhance the interpretability of image data, users are also given the option to visualize attention maps generated by the DETR model. These maps highlight spatial regions that contributed most to the model’s decision making process, aiding clinical understanding and trust in AI-based interpretations. Together, these components ensure that both tabular and image-based inferences are actionable, transparent, and aligned with clinical decision-making workflows.

\begin{figure*}[t]
  \centering
  \includegraphics[width=\textwidth]{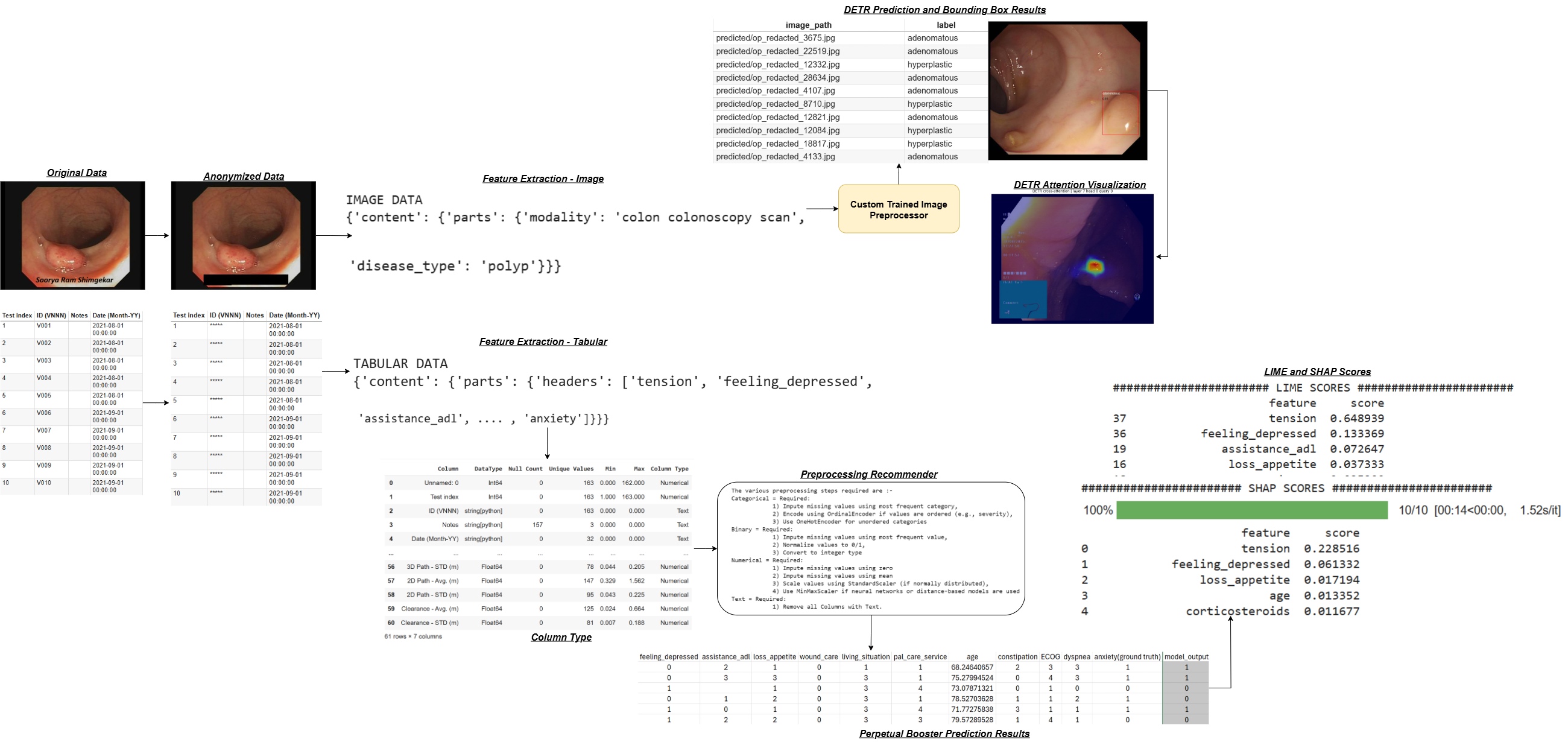}
  \caption{An example of the complete workflow for both image/tabular data}
  \label{fig:full_res}
\end{figure*}

Finally, the complete workflow is demonstrated using an example image and tabular data in \autoref{fig:full_res}.


\section{Limitations/Future Work}   
Despite the modularity and automation introduced by the agentic framework, several limitations affect its generalizability and scalability in real-world healthcare settings. One core limitation lies in the feature-model matching process, which relies on cosine similarity between SapBERT-based embeddings of user-uploaded feature names and model-required headers. While effective in many cases, this method breaks down when user features are non-standard, ambiguous, or semantically distant from known clinical terms, resulting in failed or suboptimal model selection. The preprocessing recommendation engine also presents limitations. Its current rule-based design does not incorporate learning from historical outcomes or model performance, resulting in static suggestions that may not generalize across diverse datasets or evolving model requirements. Future iterations could benefit from feedback-aware mechanisms that adapt preprocessing strategies over time based on metadata, user corrections, and inference results. Additionally, the framework assumes access to cloud-based infrastructures, particularly Google's Agent Development Kit (ADK) and related services. This assumption poses challenges for low-resource environments or institutions governed by strict data sovereignty laws, where cloud access is restricted or infeasible.

\section{Ethical Considerations and Implications}

The proposed framework currently applies data anonymization using Google Cloud’s DLP API, which provides strong support for HIPAA compliance. However, it lacks flexibility to adapt to other evolving regulations such as GDPR, PIPEDA, or local data localization laws. Future versions should include a policy-aware layer that can adjust anonymization strategies based on region-specific rules, using rule engines or compliance metadata.

Another concern is data sovereignty. Since the framework relies on cloud-based anonymization, sensitive data is temporarily sent to external servers, which may violate regulations in regions where data must remain on-site. To address this, future deployments should support privacy-preserving methods that can run locally, such as federated anonymization or edge-based redaction, ensuring data stays within secure institutional boundaries.

The use of autonomous agents for medical tasks also raises questions about accountability. As decisions are distributed across different agents, it becomes harder to trace responsibility, especially in clinical scenarios like diagnosis or risk assessment. There is a need for clear governance structures and audit trails to ensure that actions can be tracked to specific agents or supervisors.

Finally, the framework lacks formal evaluation standards. Without quantitative benchmarks and human-centered evaluations, it is difficult to assess safety, reliability, or clinical trust. Future work should include structured testing and involve domain experts to validate interpretability, usability, and ethical soundness in real-world use.

\section{Conclusion}


A key contribution of this work is the development of a feature extraction and model-data matcher agent and a preprocessing recommender, which enabled automated, modality-aware transformations tailored to specific model requirements. For tabular data, we introduce an embedding-based feature matching and model selection strategy that fundamentally aligns user uploaded data headers with model headers using semantic similarity. For image data, we develop a multi-stage MedGemma-based pipeline that first infers the imaging modality and disease category, then selects the most suitable model through vision-language reasoning. Together, these components ensure, stepwise alignment between input data and various saved models, supporting automated inferences across structured and unstructured clinical tasks. These components reduce the need for expert intervention, making the framework particularly suitable for settings where data science resources are limited. Furthermore, built-in compliance mechanisms, such as HIPAA-aligned anonymization via Google DLP, enhance its readiness for deployment in sensitive clinical environments.



In sum, this work provides a blueprint for scalable, semantically intelligent, and ethically grounded AI systems in healthcare. By embedding autonomous reasoning into each stage of the pipeline, the agentic framework holds promise for accelerating safe, interpretable, and cost-effective clinical AI adoption.

\bibliographystyle{ieeetr}
\bibliography{mybib}

\end{document}